\documentclass[11pt,a4paper]{article}
\usepackage[hyperref]{acl2021}
\usepackage{times}
\usepackage{latexsym}

\usepackage{microtype}

\usepackage{framed}

\usepackage{amsmath}

\usepackage{graphicx}

\usepackage{amssymb}% http://ctan.org/pkg/amssymb
\usepackage{pifont}% http://ctan.org/pkg/pifont
\newcommand{\cmark}{\ding{51}}%
\newcommand{\xmark}{\ding{55}}%

\usepackage{color}
\usepackage{xcolor}

\newcommand{\hide}[1]{}

\usepackage{multirow}

\usepackage{array}
\newcolumntype{L}[1]{>{\raggedright\let\newline\\\arraybackslash\hspace{0pt}}m{#1}}
\newcolumntype{C}[1]{>{\centering\let\newline\\\arraybackslash\hspace{0pt}}m{#1}}
\newcolumntype{R}[1]{>{\raggedleft\let\newline\\\arraybackslash\hspace{0pt}}m{#1}}

\def\hw{\hat{\theta}}

\def\IF{\texttt{IF}}
\def\IFp{\texttt{IF}$^+$}
\def\IFpp{\texttt{IF}$^{++}$}
\def\tracin{\texttt{TracIn}}
\def\tracinf{\texttt{TracInF}}
\def\tracinp{\texttt{TracIn$^+$}}
\def\tracinpp{\texttt{TracIn$^{++}$}}
\def\sag{\texttt{Sag}}
\def\lag{\texttt{Lag}}
\def\ral{\texttt{Ral}}
\def\cmark{\checkmark}
\aclfinalcopy

\title{On Sample Based Explanation Methods for NLP: \\
Efficiency, Faithfulness, and Semantic Evaluation}

\author{Wei Zhang \thanks{\text{  } Equal Contribution. Wei Zhang did the work while being a research scientist at IBM T.J. Watson Research Center at Yorktown Heights, NY, USA; Ziming Huang was a research scientist at IBM Research Lab at Beijing, China.} \\
  Wayfair \\
  Boston MA, USA \\
  \texttt{wzhang5@wayfair.com} \\\And
    Ziming Huang \footnotemark[1]\\
  Sogou Inc \\ Beijing, China\\
  \texttt{hzmyouxiang@gmail.com} \\\And
  Yada Zhu\\
  MIT-IBM Watson AI Lab\\ IBM Research, NY, USA \\
  \texttt{yzhu@us.ibm.com} \\\AND
  Guangnan Ye\\
  IBM Research \\ New York, USA \\
  \texttt{gye@us.ibm.com} \\\And
  Xiaodong Cui\\
  IBM Research \\ New York, USA \\
  \texttt{xcui@us.ibm.com} \\\And
  Fan Zhang \\
  IBM Data and AI \\ Littleton MA, USA\\
  \texttt{fzhang@us.ibm.com} \\}
  
\date{}

\begin{document}
\maketitle

\begin{abstract}
In the recent advances of natural language processing, the scale of the state-of-the-art models and datasets is usually extensive, which challenges the application of sample-based explanation methods in many aspects, such as explanation interpretability,  efficiency, and faithfulness. In this work, for the first time, we can improve the interpretability of explanations by allowing arbitrary text sequences as the explanation unit.
On top of this, we implement a hessian-free method with a model faithfulness guarantee. 
Finally, to compare our method with the others, we propose a semantic-based evaluation metric that can better align with humans' judgment of explanations than the widely adopted diagnostic or re-training measures.
The empirical results on multiple real data sets demonstrate the proposed method's superior performance to popular explanation techniques such as Influence Function or TracIn on semantic evaluation.

\end{abstract}

\section{Introduction}

As complex NLP models such as the Transformers family \cite{vaswani-2017-attention, devlin2018bert} become an indispensable tool in many applications, there are growing interests to explain the working mechanism of these 
``black-box'' models. Among the vast of existing techniques for explaining machine learning models, Influence Functions \cite{hampel1974influence,koh2017understanding}
%\textcolor{red}{(GN comment:we'd better mention TracIn with citation as well here)} 
that uses training instances as explanations to a model's behavior have gained popularity in NLP very recently. Different from other methods such as using input erasure \cite{li2016understanding}, saliency maps or attention matrices %\cite{serrano-smith-2019-attention,jain-wallace-2019-attention,wiegreffe-pinter-2019-attention} 
\cite{serrano-smith-2019-attention,attention_not_exp_Sarthak2019,attention_not_not_exp_Sarah} that only look at how a specific input or input sequence impacts the model decision, explaining with training instances can cast light on the knowledge a model has encoded about a problem, by answering questions like '\textit{what knowledge did the model capture from which training instances so that it makes decision in such a manner during test?}'.
Very recently, the method has been applied to explain BERT-based \cite{devlin2018bert} text classification \cite{han2020explaining,meng2020pair} and natural language inference \cite{han2020explaining} models, as well as to aid text generation for data augmentation \cite{yang2020g} using GPT-2 \cite{radford2019language}. Although useful, Influence Function may not be entirely bullet-proof for NLP applications.

First, following the original formulation \cite{koh2017understanding}, the majority of existing works use entire training instances as explanations. However, for long natural language texts that are common in many high-impact application domains (e.g., healthcare, finance, or security), it may be difficult, if not impossible, to comprehend an entire instance as an explanation. For example, a model's decision may depend only on a specific part of a long training instance.

Second, for modern NLP models and large-scale datasets, the application of Influence Functions can lead to prohibitive computing costs due to inverse Hessian matrix approximation. Although hessian-free influence score such as {\tracin} \cite{pruthi2020estimating} was introduced very recently, it may not be faithful to the model in question and can result in spurious explanations for the involvement of sub-optimal checkpoints.

Last, 
%despite the surge of efforts devoted to providing explanations to NLP models, 
the evaluation of explanation methods, in particular, for the training-instance-based ones, remains an open question. Previous evaluation is either under an over-simplified assumption on the agreement of labels between training and test instances \cite{2020Evaluation, han2020explaining} or is based on indirect or manual inspection \cite{hooker2018benchmark,meng2020pair,han2020explaining,pruthi2020evaluating}. A method to automatically measure the semantic relations at scale and that highly correlates to human judgment is still missing in the evaluation toolset.

To address the above problems, we propose a framework to explain model behavior that includes both a set of new methods and a new metric that can measure the semantic relations between the test instance and its explanations. The new method allows for arbitrary text spans as the explanation unit and is Hessian-free while being faithful to the final model. Our contributions are: 
\begin{enumerate}
    \item We propose a new explanation framework that can use arbitrary explanation units as explanations and be Hessian-free and faithful at the same time;
    \item A new metric to measure the semantic relatedness between a test instance and its explanation for BERT-based deep models.
\end{enumerate}

%The rest of the paper is organized as follows: we first introduce the preliminaries of existing methods, then introduce our method, followed by details on the data, model, and implementation details of the proposed explanation method. Finally, we compare explanation methods on the set of metrics, including our new evaluation metric, followed by the analysis and conclusions. 

\section{Preliminaries}
Suppose a model parameterized by $\hat\theta$ is trained on classification dataset $D=\{D^{train}, D^{test}\}$ by empirical risk minimization over $D^{train}$. Let $z=(\mathbf{x},y) \in D^{train}$ and $z'=(\mathbf{x}',y') \in D^{test}$ denote a training and a test instance respectively, where $\mathbf{x}$ is a token sequence, and $y$ is a scalar. The goal of training instance based explanation is to provide for a given test $z'$ an ordered list of training instances as explanation. Two notable methods to calculate the influence score are {\IF} and {\tracin}:

\textbf{\IF} \cite{koh2017understanding}  assumes the influence of $z$ can be measured by perturbing the loss function $L$ with a fraction of the loss on $z$, and obtain
\begin{align}
\label{eq:IF}
    \begin{split}
    \mathcal{I}_{\text{pert,loss}} & (z, z' ; \hat{\theta})  \\
    &= - \nabla_{{\theta}} L(z', \hat{\theta}) H^{-1}_{\hat{\theta}} \nabla_{\theta} L(z, \hat{\theta}),
    \end{split}
\end{align}
where $H$ is the Hessian matrix calculated on the entire training dataset, a potential computation bottleneck for large dataset $D$ and complex model with high dimensional $\hw$.

\textbf{\tracin} \cite{pruthi2020estimating} instead assumes the influence of a training instance $z$ is the sum of its contribution to the overall loss all through the entire training history, and conveniently it leads to 
\begin{align}
\label{eq:tracin}
    \begin{split}
    \texttt{TracIn}(z,z') & =  \\
    \sum_i \eta_i \nabla_{ \hat \theta_i} &  L(\hw_i , z) \nabla_{ \hat \theta_i} L(\hw_i , z'),
        \end{split}
\end{align}
where $i$ iterates through the checkpoints saved at different training steps and $\eta_i$ is a weight for each checkpoint. {\tracin} does not involve Hessian matrix and more efficient to compute. We can summarize the key differences between them according to the following desiderata of an explanation method:
\paragraph{Efficiency}for each $z'$, {\tracin} requires $\mathcal{O}(CG)$ where $C$ is the number of models and $G$ is the time spent for gradient calculation; whereas \IF{} needs $\mathcal{O}(N^2G)$ where $N$ is the number of training instances, and $N>>C$ in general. \footnote{some approximation such as hessian-inverse-vector-product \cite{baydin2016tricks} may improve efficiency to $O(NSG)$ where $S$ is the approximation step and $S<N$} 

\paragraph{Faithfulness} \IF{} is \textit{faithful} to $\hat\theta$ since all its calculation is based on a single final model, yet {\tracin} may be less faithful to $\hat\theta$ since it obtains gradients from a set of checkpoints \footnote{We may say {\tracin} is faithful to the \textit{data} rather than to the \textit{model}. And in the case where checkpoint averaging can be used as model prediction, the number of checkpoints may be too few to justify Eq. \ref{eq:tracin}.}. 

\paragraph{Interpretability} Both methods use the entire training instance as an explanation. Explanations with a finer-grained unit, e.g., phrases, may be easier to interpret in many applications where the texts are lengthy.

\section{Proposed Method}
To improve on the above desiderata, a new method should be able to: 1) use any appropriate granularity of span(s) as the explanation unit; 2) avoid the need of Hessian while maintaining faithfulness. We discuss the solutions for both in Section \ref{subsec:improve_interp} and \ref{subsec:faithful_expl}, and combine them into one formation in Section \ref{subsec:combined} followed by critical implementation details.

\subsection{Improved Interpretability with Spans}
\label{subsec:improve_interp}
To achieve 1), we first start with influence functions \cite{koh2017understanding} and consider an arbitrary span of training sequence $\mathbf{x}$ to be evaluated for the qualification as explanation \footnote{the method can be trivially generalized to multiple spans}. Our core idea is \textit{to see how the model loss on test instance $z'$ changes with the training span's importance.} The more important a training span is to $z'$, the greater this influence score should be. We derive it in three following steps.

First, we define the training span from token $i$ to token $j$ to be $\mathbf{x}_{ij}$, and the sequence with $\mathbf{x}_{ij}$ masked is $\mathbf{x}_{-ij}=[x_0, ..., x_{i-1}, \text{[MASK]}, ..., \text{[MASK]},  x_{j+1},...]$ and its corresponding training data is $z_{-ij}$. We use logit difference \cite{li2020bert} as importance score based on
%of the meaning unit $x_{kl}$ of test data $z'$ 
the empirical-risk-estimated parameter $\hat \theta$ obtained from $D^{train}$ as: $    \texttt{imp} (\mathbf{x}_{ij} | z, \hat{\theta} ) = \text{logit}_y(\mathbf{x}; \hat{\theta}) - \text{logit}_y(\mathbf{x}_{-ij}; \hat{\theta})$, where every term in the right hand side (RHS) is the logit output evaluated at a model prediction $y$ from model $\hat\theta$ right before applying the SoftMax function. \textit{This equation tells us how important a training span is.} It is equivalent to the loss difference 
\begin{equation}
\label{eq:test_imp2}
    \texttt{imp}(\mathbf{x}_{ij}| z ; \hat{\theta} ) = \mathcal{L}( z_{-ij};\hat{\theta} ) -  \mathcal{L} ( z;\hat{\theta} ),
\end{equation}
when the cross entropy loss $\mathcal{L}( z;\theta )= - \sum_{y_i} \mathcal{I}(y=y_i) \text{logit}_{y_i} (\mathbf{x};\theta)$ is applied.

Then, we measure $\mathbf{x}_{ij}$'s influence on model $\hat\theta$ by adding a fraction of \texttt{imp}($\mathbf{x}_{ij}| z ; \hat{\theta}$) scaled by a small value $\epsilon$ to the overall loss and obtain $\hat\theta_{\epsilon, \mathbf{x}_{ij}|z } := \text{argmin}_\theta E_{z_i\in D^{train}}[\mathcal{L}(z_i,\theta)] + \epsilon \mathcal{L}(z_{-ij};\theta) - \epsilon \mathcal{L}(z;\theta)$. Applying the classical result in \cite{cook1982residuals,koh2017understanding}, the influence of up-weighing the importance of $x_{ij}$ on $\hat\theta$ is
\begin{multline*}
\frac{d \hat{\theta}_{\epsilon, \mathbf{x}_{ij}|z}} {d\epsilon} \Big |_{\epsilon=0}= \\ 
H_{\hat{\theta}}^{-1} (\nabla_{\hat \theta}L(z; \hat{\theta})  - \nabla_{\hat \theta} L(z_{-ij};\hat{\theta})).
\end{multline*}

Finally, applying the above equation and the chain rule, we obtain the influence of $\mathbf{x}_{ij}$ to $z'$ as:
\begin{align*}
%\label{eq:IFA}
     & \IF^+(\mathbf{x}_{ij}| z, z' ; \hat\theta) := \nabla_{\epsilon} L( z' ; \hat\theta_{\epsilon, \mathbf{x}_{ij}|z} ) |_{\epsilon=0} \nonumber \\
    &  = \nabla_{\theta} L (z'; \hat\theta)  H_{\hat\theta}^{-1}(\nabla_{\theta} L(z; \hat\theta) -\nabla_{\theta} L(z_{-ij}; \hat\theta) ) .
\end{align*}

%\IFp obtained from Eq. \ref{eq:IFA} provides a principled way to measure how the change of importance of any span/spans in a training instance $z$ influence a test instance $z'$. 
\IFp{} measures the influence of a training span on an entire test sequence.
Similarly, we also measure the influence of a training span to a test span $x'_{kl}$ by applying Eq. \ref{eq:test_imp2} and obtain 
\begin{align*}
%\label{eq:IFpp}
    \begin{split}
     & \IF^{++}(\mathbf{x}_{ij}| z, \mathbf{x}'_{kl}|z'; \hat\theta)\\ := & \nabla_{\epsilon} L(z_{-kl}'; \hat\theta_{\epsilon, \mathbf{x}_{ij}|z} ) - \nabla_{\epsilon} L(z'; \hat\theta_{\epsilon, \mathbf{x}_{ij}|z} ) |_{\epsilon=0} \\
    = & (\nabla_{\theta} L (z'_{-kl}; \hat\theta) -\nabla_{\theta} L (z'; \hat\theta) ) \\
    & H_{\hat\theta}^{-1}(\nabla_{\theta} L(z; \hat\theta) -\nabla_{\theta} L(z_{-ij}; \hat\theta) ) .
    \end{split}
\end{align*}
The complete derivation can be found in Appendix.

\paragraph{On the choice of Spans} Theoretically, \IFp{} and \IFpp{} can be applied to any text classification problem and dataset with an appropriate choice of the span. If no information about valid span is available, shallow parsing tools or sentence split-tools can be used to shatter an entire text sequence into chunks, and each chunk can be used as span candidates. In this situation, the algorithm can work in two steps: 1) using masking method \cite{li2020bert} to determine the important test spans; and 2) for each span we apply \IFpp{} to find training instances/spans as explanations.  

Usually, we can choose top-K test spans, and even can choose K$=$1 in some cases. In this work, we look at the later case without loss of generality, and adopt two aspect-based sentiment analysis datasets that can conveniently identify a deterministic span in each text sequence, and frame the span selection task as a Reading Comprehension task \cite{rajpurkar-etal-2016-squad}. We discuss the details in Section \ref{sec:data}. Note that the discussion can be trivially generalized to the case where K$>$1 using Bayesian approach such as $\text{imp}(x_{ij})=\mathbf{E}_{P(x_{kl}')} [\text{imp}(x_{ij}|x_{kl})']$ which can be explored in future work.

\subsection{Faithful \& Hessian-free Explanations}
\label{subsec:faithful_expl}
To achieve 2), we would start with the method of {\tracin} \cite{pruthi2020estimating} described in Eq. \ref{eq:tracin} which is Hessian free by design. {\tracin} defines the contribution of a training instance to be the sum of its contribution (loss) throughout the entire training life cycle, which eradicated the need for Hessian. However, this assumption is drastically different from \IF{}'s where the contribution of $z$ is obtained solely from the final model $\hat\theta$. By nature, \IF{} is a \textit{faithful method}, and its explanation is \textit{faithful to} $\hat\theta$, and \tracin{} in its vanilla form is arguably not a faithful method.

\paragraph{Proposed treatment} Based on the assumption that the influence of $z$ on $\hw$ is the sum of influences of all variants close to $\hw$, we define a set of ``faithful'' variants satisfying the constraint of $\{ \hat\theta_i| 1> \delta >> ||\hat\theta_i-\hat\theta||_2\}$, namely $\delta$-faithful to $\hat\theta$. The smaller $\delta$ is, the more faithful the explanation method is. Instead, the $\delta$ for {\tracin} can be arbitrary large without faithfulness guarantees, as some checkpoints can be far from the final $\hw$. Thus, we construct a $\delta$-faithful explanation method that mirrors {\tracin} as: 
\begin{align*}
%\label{eq:tracinf}
    \begin{split}
    \texttt{TracInF}(z,z') & =  \\
    \sum_i \nabla_{ \hw +\delta_i} &  L(\hw + \delta_i , z) \nabla_{ \hw+\delta_i } L(\hw + \delta_i , z').
        \end{split}
\end{align*}
The difference between {\tracin} and \texttt{TracInF} is that the checkpoints used in {\tracin} are correlated in time whereas all variants of \texttt{TracInF} are conditionally independent. 
Finding a proper $\delta_i$ can be tricky. If ill-chosen, $\delta_i$ may diverge $\hw$ so much that hurts gradient estimation. In practice, we estimate $\delta_i=\eta_i g(z_i|\hw)$ obtained from a single-step gradient descent $g(z_i|\hw)$ with some training instance $z_i$ on model $\hw$, scaled by an $i$-specific weighting parameter $\eta_i$, which in the simplest case is uniform for all $i$. Usually $\eta_i$ should be small enough so that $\hw+\delta_i$ can stay close to $\hw$. In this paper we set $\eta$ as the model learning rate for proof of concept.

\paragraph{Is \tracin{\texttt{F}} faithful?} First, any $\hw+\delta_i$ is close to $\hw$. Under the assumption of Lipschitz continuity, there exists a $k\in \mathbb{R}^+$  such that $\nabla L(\hw+\delta_i, z)$ is bounded around $\nabla L(\hw, z)$ by $k |\eta_i g^2(z_i|\hw)|$, the second derivative, because $|\nabla L(\hw+\delta_i, z)-\nabla L(\hw, z)|<k|\eta_i g^2(z_i|\hw)|$. A proper $\eta_i$ can be chosen so that the right hand side (RHS) is sufficiently small to bound the loss within a small range. Thus, the gradient of loss, and in turn the \tracin{\texttt{F}} score can stay $\delta$-faithful to $\hw$ for an sufficiently small $\delta$, which \tracin{} can not guarantee.

\subsection{The Combined Method}
\label{subsec:combined}
By combining the insights from Section \ref{subsec:improve_interp} and \ref{subsec:faithful_expl}, we obtain a final form named \tracinpp{}:
\begin{align*}
%\label{eq:powertracin}
\begin{split}
& \texttt{TracIn}^{++}  (x'_{kl}|z' , x_{ij}| z ; \hw)  = \\
&\sum_i \big [ \nabla \mathcal{L}(\hw+\delta_i, z'_{-kl}) -\nabla \mathcal{L}(\hw+\delta_i, z') \big]  \\
&\big[\nabla \mathcal{L}(\hw+\delta_i, z) - \nabla \mathcal{L}(\hw+\delta_i, z_{-ij}) \big].
\end{split}
\end{align*}
This ultimate form mirrors the \IFpp{} method, and it satisfies all of our desiderata on an improved explanability method. Similarly, \tracinp{} that mirrors \IFp{} is
\begin{align*}
%\label{eq:tracinp}
\begin{split}
& \texttt{TracIn}^{+}  ( z' , x_{ij}| z ; \hw)  = \sum_i  \nabla \mathcal{L}(z'; \hw+\delta_i)  \\
&\big[\nabla \mathcal{L}(\hw+\delta_i, z) - \nabla \mathcal{L}(\hw+\delta_i, z_{-ij}) \big].
\end{split}
\end{align*}

\subsection{Additional Details}
Since the RHS of \IF{}, \IFp{} and \IFpp{} equations all involve the inverse of Hessian Matrix, here we discuss the computation challenge. Following \cite{koh2017understanding}, we adopt the vector-Hessian-inverse-product (VHP) with stochastic estimation \cite{baydin2016tricks}. The series of stochastic updates, one for each training instance, is performed by the vhp() function in the \texttt{torch.autograd.functional} package and the update stops until convergence.  
%We first calculate the product of the first two terms on the RHS, i.e., the vector-Hessian-inverse-product (VHP), using a stochastic approach \cite{baydin2016tricks} that approximates the product with a series of stochastic updates, one for each training instance, a fashion similar to SGD. Each update is performed by the vhp() function in the \texttt{torch.autograd.functional} package and the update stops until convergence.
%(Algorithm \ref{algo:vhp})
Unfortunately, we found that naively applying this approach leads to VHP explosion due to large parameter size. To be specific, in our case, the parameters are the last two layers of \texttt{RoBERTa-large} \cite{roberta-large} plus the output head, a total of ~12M parameters per gradient vector. To stabilize the process, we take three approaches: 1) applying gradient clipping (set to 100) to avoid accumulating the extreme gradient values; 2) adopting early termination when the norm of VHP stabilizes (usually $<1000$ training instances, i.e., the depth); and 3) slowly decaying the accumulated VHP with a factor of 0.99 (i.e., the damp) and update with a new vhp() estimate with a small learning rate (i.e., the scale) of 0.004. Please refer to our code for more details. Once obtained, the VHP is first cached and then retrieved to perform the dot-product with the last term. The complexity for each test instance is $\mathcal{O}(dt)$ where $d$ is the depth of estimation and $t$ is the time spent on each vhp() operation. The time complexity of different \IF{}  methods only vary on a constant factor of two.

For each of \tracin{}, \tracinp{} and \tracinpp{}, we need to create multiple model variants. For \tracin{}, we save three checkpoints of the most recent training epochs; For \tracinp{} or \tracinpp{}, we start with the same checkpoint and randomly sample a mini-batch 3 times and perform one-step training (learning rate  1E-4) for each selection to obtain three variants. We do not over-tune those hyper-parameters for replicability concerns.

\section{Evaluation Metrics}
This section introduces our semantic evaluation method, followed by a description of two other popular metrics for comparison.

\subsection{Semantic Agreement (\sag{})} 
Intuitively, a rational explanation method should rank explanations that are semantically related to the given test instance relatively higher than the less relevant ones. \textit{Our idea is to first define the semantic representation of a training span $\mathbf{x}_{ij}$ of $z$ and measure its similarity to that of a test span $\mathbf{x}'_{kl}$ of $z'$}. 
%\zhu{To Wei, it is better to add the meaning of $x_ij$ and z here to increase readability, as here is far from their definition. reviewer could forget their meaning and get lost. } 
Since our method uses \texttt{BERT} family as the base model, we obtain the embedding of a training span by the difference of $x$ and its span-masked version $x_{ij}$ as
\begin{equation}
\label{eq:span_emb}
    emb(x_{ij}) = emb(x) - emb(x_{-ij}),
\end{equation}
where $emb$ is obtained from the embedding of sentence start token such as ``[CLS]'' in BERT \cite{devlin2018bert} at the last embedding layer. To obtain embedding of the entire sequence we can simply use the $emb(x)$ without the last term in Eq. \ref{eq:span_emb}. Thus, all spans are embedded in the same semantic space and the geometric quantities such as cosine or dot-product can measure the similarities of embeddings.
%, an assumption behind many NLP models \cite{wang2018glue,wang2019superglue,rajpurkar-etal-2016-squad}. 
We define the semantic agreement \texttt{Sag} as:
\begin{equation}
\begin{split}
    &  \sag{}(z', \{z\}|_1^K) = \\
    &\frac{1}{K}\sum_{z} cos(emb(x_{ij}|z), emb(x'_{kl}|z')),
\end{split}
\end{equation}
Intuitively, the metric measures the degree to which top-K training spans align with a test span on semantics.
%We use the mean and variance of \texttt{Sag} over many test instances $z'$ to compare explanation methods.

\subsection{Other metrics}
\paragraph{Label Agreement (\lag{})} label agreement \cite{2020Evaluation} assumes that the label of an explanation $z$ should agree with that of the text case $z'$. Accordingly, we retrieve the top-K training instances from the ordered explanation list and calculate the label agreement (\texttt{Lag}) as follows:
\begin{equation*}
\begin{split}
    \lag{}(z', \{z\}|_1^N)=&\frac{1}{K}\sum_{k\in[1,K]} \mathcal{I}(y'==y_k),\\
%    + \frac{1}{2K} & \sum_{k\in[N-K,N]} \mathcal{I}(y'<>y_k)
\end{split}
\end{equation*}
where $\mathcal{I}(\cdot)$ is an indicator function. \lag{} measures the degree to which the top-ranked $z$ agree with $z'$ on class label, e.g., if the sentiment of the test $z'$ and explanation $z$ agree.
        
\paragraph{Re-training Accuracy Loss (\ral{})}
\ral{} measures the loss of test accuracy after removing the top-K most influential explanations identified by an explanation method \cite{2020Evaluation, hooker2018benchmark, han2020explaining}. The assumption is that the higher the loss the better the explanation method is. Formally, 
\begin{equation*}
    \ral{}(f,\hw)= Acc(\hw) - Acc(\hw'),
\end{equation*}
where $\hw'$ is the model re-trained by the set $D^{train}/\{z\}|_1^K$. Notice the re-training uses the same set of hyper-parameter settings as training (Section \ref{sec:train}). To obtain $\{z\}|_1^K$, we combine the explanation lists for all test instances (by score addition) and then remove the top-K from this list.  

\section{Data}
\label{sec:data}
Our criteria for dataset selection are two folds: 1. The dataset should have relatively high classification accuracy so that the trained model can behave rationally; and 2. The dataset should allow for easy identification of critical/useful text spans to compare span-based explanation methods. We chose two aspect-based sentiment analysis (ABSA) datasets; one is ATSA, a subset of \textbf{MAMS} \cite{jiang2019challenge} for product reviews, where aspects are the terms in the text. The other is \textbf{sentihood} \cite{saeidi2016sentihood} of location reviews. We can identify the relevant span of an aspect term semi-automatically and train models with high classification accuracy in both datasets. (see Section \ref{sec:train} for details). Data statistics and instances are in Table \ref{tab:data} and \ref{tab:data_instance}.

\begin{table}[!htbp]
\centering
\begin{tabular}{r|rrr}
%   \multicolumn{2}{c}{ } & \multicolumn{3}{c}{Data Split}  & \multicolumn{5}{c}{Label Types} \\
 & Train & Dev & Test \\
\hline
MAMS & 11186 & 1332 & 1336  \\
%\hspace{5mm}-ACSA (Cat.) & & 7090 & 888 & 901& & & \cmark & \\
sentihood & 2977 & 747 & 1491 \\
%SigmaLaw &  Legal & 1400 & 200 & 404 &\cmark & & & & \cmark \\
%PerSenT &  Politics  & 3355 & 577 & 1406 & & & & \cmark & \cmark\\
\end{tabular}
\caption{Data Statistics. Note that we regard each training instance as aspect-specific, i.e., the concatenation of aspect term and the text $x$ as model input.}
\label{tab:data}
\end{table}

\begin{table*}[!htbp]
    \centering
    \begin{tabular}{l|c|c|c}
    Dataset & Text & Aspect& Sentiment \\
    \hline
    \hline
    \multirow{3}{2cm}{MAMS} & \multirow{3}{*}{\parbox{9cm}{ the \textit{service} \textcolor{red}{was impeccable}, the \textit{menu} \textcolor{brown}{traditional but inventive} and presentation for the most part excellent but the \textit{food} itself \textcolor{blue}{came up short}.}} & \textcolor{red}{service} & + \\
%    \cline{3-4}
    & & \textcolor{brown}{menu} & + \\
%    \cline{3-4}
    & & \textcolor{blue}{food} & - \\
    \hline
        \multirow{2}{*}{sentihood} & \multirow{2}{*}{\parbox{9cm}{ i live in location2  and \textcolor{blue}{i love it} location1 just \textcolor{red}{stay away from} location1 lol.}} & \textcolor{red}{location1} & - \\
%    \cline{3-4}
    & & \textcolor{blue}{location2} & + \\
    \end{tabular}
    \caption{Dataset instances. In text, each aspect has a supporting span which we annotate semi-automatically. We choose a subset where test instances }
    \label{tab:data_instance}
\end{table*}

\paragraph{Automatic Span Annotation}
 As shown in the colored text in Table \ref{tab:data_instance}, we extract the spans for each term to serve as explanation units for \IFp{}, \IFpp{}, \tracinp{} and \tracinpp{}. To reduce annotation effort, we convert span extraction into a question answering task \cite{rajpurkar-etal-2016-squad} where we use aspect terms to formulate questions such as ``How is the \textit{service}?'' which concatenates with the text before being fed into pre-trained machine reading comprehension (RC) models. The output answer is used as the span.
When the RC model fails, we use heuristics to extract words before and after the term word, up to the closest sentence boundary. See appendix for more details. We sampled a subset of 100 annotations and found that the RC model has about 70\% of Exact Match \cite{rajpurkar-etal-2016-squad} and the overall annotation has a high recall of over 90\% but low EM due to the involvement of heuristics.

\paragraph{(Not) Mitigating the Annotation Error} Wrongly-annotated spans may confuse the explanation methods. For example, as shown in \ref{tab:data_instance}, if the span of \textit{location2} is annotated as ``I love it'', span-based explanation methods will use it to find wrong examples for explanation. Thus test instances with incorrectly annotated spans are omitted, i.e., no tolerance to annotation error for test instances. To the contrary, for training instances, we do not correct the annotation error. The major reason is the explanation methods have a chance to rank the wrongly annotated spans lower (its importance score \texttt{imp}() of Eq. \ref{eq:test_imp2} can be lower and in turn for its influence scores.) Also, It is labor-intensive to do so.

\section{Experiments}

\subsection{Model Training Details}
\label{sec:train}
%\zhu{I think the training details can be put in the appendix}
We train two separate models for MAMS and sentihood. 
The model's input is the concatenation of the aspect term and the entire text, and the output is a sentiment label.
The two models share similar settings: 1. they both use \textsc{roBERTa-large} \cite{roberta-large} from Huggingface \cite{Wolf2019HuggingFacesTS} which is fed into the \texttt{BertForSequenceClassification} function for initialization. We fine-tune the parameters of the last two layers and the output head using a batch size of 200 for ATSA and 100 for sentihood and max epochs of 100. We use AdamW optimizer \cite{loshchilov2017decoupled} with weight decay 0.01 and learning rate 1E-4. Both models are written in Pytorch and are trained on a single Tesla V100 GPU and took less than 2 hours for each model to train. The models are selected on dev set performance, and both trained models are state-of-the-art: 88.3\% on MAMS and 97.6\% for sentihood at the time of writing.

\label{sub:exp}

\begin{table*}[htb]
    \centering
    \begin{tabular}{ll|lll|lll}
& & \IF{} & \IFp{} & \IFpp{} & \tracinf{} & \tracinp{} & \tracinpp{} \\
\hline
\multicolumn{2}{c|}{\textbf{Faithful to} $\hw$?} & \cmark & \cmark & \cmark & \cmark & \cmark & \cmark \\ 
\multicolumn{2}{c|}{\textbf{Hessian-free}?} & \xmark & \xmark & \xmark & \cmark & \cmark& \cmark \\
\multicolumn{2}{c|}{\textbf{Interpretable explanations}?} & \xmark & \cmark & \cmark\cmark & \xmark & \cmark & \cmark\cmark \\
\hline
\hline
%\multirow{3}{2cm}{MAMS}

%\multicolumn{1}{l|}{\sag{}(10)} &14.22/00.35&17.17/00.96&21.74/01.18& 15.89/00.36&22.65/01.56&\textbf{23.92}/01.09\\
%\multicolumn{1}{l|}{\sag{}(100)} &14.65/00.21&15.10/00.51&19.83/07.45& 15.97/02.73&19.54/09.08&\textbf{21.32}/00.76\\
%\multicolumn{1}{l|}{\lag{}(10)} & 21.63/07.16 & 25.66/07.82 & 65.41/13.45 & 38.20/06.26&08.60/02.99 & \textbf{78.03}/10.65 \\
%\multicolumn{1}{l|}{\lag{}(100)} & 26.07/06.54 & 25.66/07.06 & 62.52/10.85 & 43.19/04.95&06.27/01.34 & \textbf{75.02}/08.45 \\

\multirow{6}{*}{MAMS} & \multicolumn{1}{l|}{\sag{}(K=10)} &14.22 &17.17 &21.74 & 15.89 &22.65 &\textbf{23.92} \\
& \multicolumn{1}{l|}{\sag{}(K=100)} &14.65 &15.10 &19.83 & 15.97 &19.54 &\textbf{21.32}\\
& \multicolumn{1}{l|}{\lag{}(K=10)} & 21.63 & 25.66 & 65.41 & 38.20 &08.60 & \textbf{78.03} \\
& \multicolumn{1}{l|}{\lag{}(K=100)} & 26.07  & 25.66  & 62.52  & 43.19 &06.27  & \textbf{75.02} \\

& \multicolumn{1}{l|}{\ral{}(- top 20\%)} & 09.80 & 05.64 & 03.55 & 09.80&11.89& \textbf{16.05} \\
& \multicolumn{1}{l|}{\ral{}(- top 50\%)} & \textbf{28.55} & 01.47 & 18.14 &22.30 & 05.64 & 18.14 \\
\hline
%\multicolumn{1}{l|}{\sag{}(10)} &04.69/02.80&04.75/08.25&22.54/10.79&03.07/02.92&00.98/07.31&\textbf{26.21}/11.30\\
%\multicolumn{1}{l|}{\sag{}(50)} &03.56/01.72&07.82/07.49&22.21/9.14&01.78/01.74&01.61/05.98&\textbf{23.43}/10.25\\
%\multicolumn{1}{l|}{\lag{}(10)} &53.00/12.37&41.91/18.66& 61.96/19.51& 55.91/13.76& 18.22/12.80&\textbf{66.65}/20.42\\
%\multicolumn{1}{l|}{\lag{}(50)} &56.38/07.98&44.05/17.67& 63.16/17.84& 59.66/06.38&17.49/10.86&\textbf{66.72}/19.34\\
\multirow{6}{*}{sentihood}& \multicolumn{1}{l|}{\sag{}(K=10)} &04.69&04.75&22.54&03.07&00.98&\textbf{26.21}\\
& \multicolumn{1}{l|}{\sag{}(K=50)} &03.56&07.82&22.21&01.78&01.61&\textbf{23.43}\\
& \multicolumn{1}{l|}{\lag{}(K=10)} &53.00&41.91& 61.96& 55.91& 18.22&\textbf{66.65}\\
& \multicolumn{1}{l|}{\lag{}(K=50)} &56.38&44.05& 63.16& 59.66&17.49&\textbf{66.72}\\

& \multicolumn{1}{l|}{\ral{}(- top 20\%)} & 10.56 &\textbf{16.21}&06.91&09.23&06.91 & 09.23\\
& \multicolumn{1}{l|}{\ral{}(- top 50\%)} & 16.21 & 18.53 & 11.05 & \textbf{27.83} & 9.23 & 4.58 \\

\end{tabular}
    \caption{Performance of difference explanation methods on 200 test cases on each dataset. For \sag{} and \lag{} we set $K \in \{10, 100\}$; for \ral{} we set $K \in \{20\%, 50\%\}$, and \ral{} we consider removing the top 20\% or 50\% from the ordered training instance list. Computation time for \IF{} family is about ~20 minutes per test instance with recursion depth 1000 (the minimal value to guarantee convergence) on a Tesla V100 GPU. The time for \tracin{} family only depends on gradient calculation, which is trivial compared to \IF{} family. }
    \label{tab:bigtable}
\end{table*}

\subsection{Comparing Explanation Methods}
We compare the six explanation methods on two datasets and three evaluation metrics in Table \ref{tab:bigtable} from which we can draw the following conclusions:

1) \tracin{} family outperforms \IF{} family according to \sag{} and \lag{} metrics. We see that both metrics are robust against the choice of $K$. It it worth noting that \tracin{}  family methods are not only efficient, but also effective for extracting explanations compared to \IF{} family as per \sag{} and \lag{}.

2) Span-based methods (with +) outperform Vanilla methods (w/o +). It is good news because an explanation can be much easier to comprehend if we can highlight essential spans in text, and \IFpp{} and \tracinpp{} shows us that such highlighting can be justified by their superiority on the evaluation of \sag{} and \lag{}.

3) \sag{} and \lag{} shows a consistent trend of \tracinpp{} and \IFpp{} being superior to the rest of the methods, while \ral{} results are inconclusive, which resonates with the findings in \cite{hooker2018benchmark} where they also observed randomness after removing examples under different explanation methods. This suggests that the re-training method may not be a reliable metric due to the randomness and intricate details involved in the re-training process.

4) The \sag{} measures \tracinp{} differently than \lag{} shows that \lag{} may be an over-simplistic measure by assuming that label $y$ can represent the entire semantics of $\mathbf{x}$, which may be problematic. But \sag{} looks into the $\mathbf{x}$ for semantics and can properly reflect and align with humans judgments. 

\begin{figure}[htb]
    \centering
   \includegraphics[width=0.48\textwidth, angle=0]{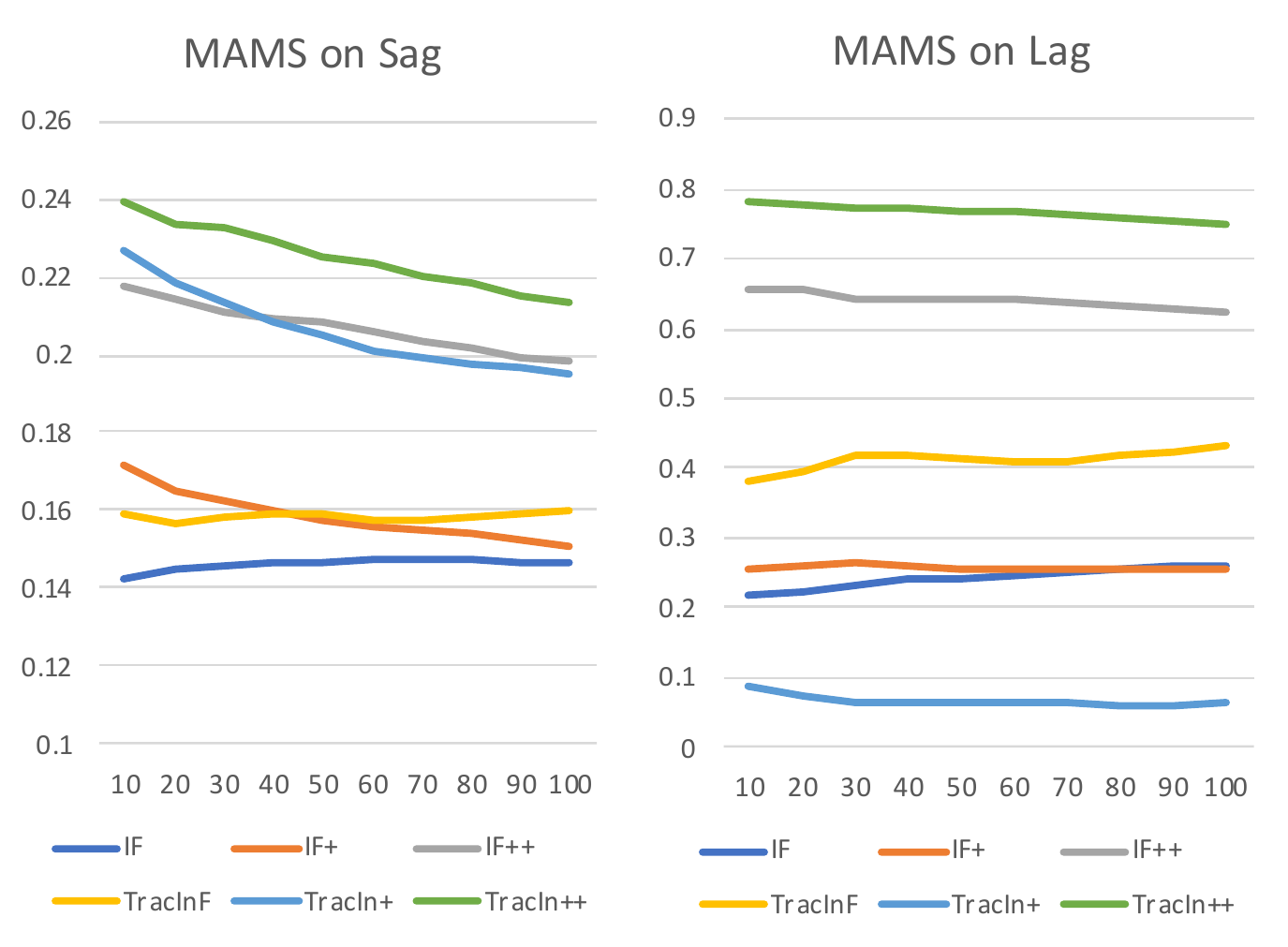}    \caption{ \sag{} and \lag{} v.s. K values on 200 MAMS test instances.}
    \label{fig:k_trend}
\end{figure}

\paragraph{The Impact of K on Metrics}
One critical parameter for evaluation metrics is the choice of K for \sag{} and \lag{} (We do not discuss K for \ral{} due to its randomness). Here we use 200 MAMS test instances as subjects to study the influence of K, as shown in Figure \ref{fig:k_trend}. 

We found that as K increases, all methods, except for \IF{} and \tracinf{}, decrease on \sag{} and \lag{}. The decrease is favorable because the explanation method is putting useful training instances before less useful ones. In contrast, the increase suggests the explanation method fails to rank useful ones on top. This again confirms that span-based explanation can take into account the useful information in $\mathbf{x}$ and reduce the impact of noisy information involved in \IF{} and \tracinf{}.

\subsection{Comparing Faithfulness}
How faithful our proposed \tracinpp{} to $\hw$? To answer this question, we first define the notion of \textit{strictly faithful explanation} and then test an explanation method's faithfulness against it. Note that none of the discussed methods is strictly faithful, since \IFpp{} used approximated inverse-Hessian and \tracinpp{} is a $\delta$ away from being strictly faithful. To obtain ground truth, we modify \tracinpp{} to use a single checkpoint $\hw$ as the ``ultimately faithful'' explanation method \footnote{The choice of ground truth can also be the exact computation of inverse-Hessian in \IF{} (our future work). Faithfulness does not equal to correctness; there is no guarantee the ground truth is a valid explanation method, but it can be a valid benchmark for faithfulness}. Then, we obtain an explanation list for each test instance and compute its Spearman Correlation with the list obtained from the ground truth. The higher the correlation, the more faithful the method is.

\begin{table}[htb]
    \centering
    \begin{tabular}{c|cc}
        & \multicolumn{2}{c}{Spearman} \\
        Method &  Mean &  Var.  \\
        \hline\hline
        Control & 55.11 & 4.84 \\
        \hline
        \tracinpp{} & 60.14 & 3.57 \\%& 48.15 (2.16) \\
        \hline
        \IFpp{} & 59.37 & 20.50  \\% & 62.81 (22.73)  \\
    \end{tabular}
    \caption{Comparison of Correlation with Ground truth. The experiment is run 5 times each; ``Control'' is only different from \tracinpp{} on the models used: ``control'' uses three checkpoints of the latest epochs, but \tracinpp{} uses three $\delta$-faithful model variants.}
    \label{tab:faithfulness}
\end{table}
 
In Table \ref{tab:faithfulness} we discovered that \tracinpp{} has similar mean as \IFpp{} but has a much lower variance, showing its stability over \IFpp{}. This aligns with the finding of Basu et al. \shortcite{basu2020influence} which argues that in deep non-convex networks, influence function usually is non-stable across test instances. \tracin{} family arguably may be a promising direction to stability. Both methods are more faithful to Ground truth than Control that uses checkpoints, showing that the model ``ensemble'' around $\hw$ may be a better choice than ``checkpoint averaging'' for model explanations. Further explorations may be needed since there are many variables in this comparison.

\section{A Case Study}
Table \ref{tab:cases} demonstrate the differences of explanation methods. In action, \tracinpp{} shows both the test span and explanation span to a user; \tracinp{} shows only the training span, and \tracin{} does not show spans. Interestingly we can observe the top-1 explanation found by \tracinpp{} is more semantically related than others in the example, a common pattern among the test cases.

\begin{table*}[]
    \centering
    \begin{tabular}{L{1.8cm}|L{12cm}|C{0.9cm}}
%        & & senti. \\
        \hline
        \textbf{Test Case} & been here a few times and \textcolor{blue}{food} \textcolor{gray}{has always been good} but service really suffers when it gets crowded. & + \\
        \hline\hline
        \tracinpp{} & expected there to be more options for tapas the food was mediocre but the \textcolor{blue}{service} \textbf{was pretty good.} & + \\
        \hline
        \tracinp{} & \textcolor{blue}{deco}r \textbf{is simple yet functional} and although the staff are not the most attentive in the world, ... & + \\
        \hline
        \tracinf{} & this place is the tourist fav of \textcolor{blue}{chinese food} in the city, the service was fast, but the taste of the food is average, too much starch ... & 0 \\
        \hline\hline
        \IFpp{} & ... the host was rude to us as we walked in, we stayed because the \textcolor{blue}{decor} is \textbf{charming} and we wanted french food. & + \\
        \hline
        \IFp{} & the scene \textbf{a dark refurbished} \textcolor{blue}{dining car} hosts plenty of hipsters in carefully selected thrift-store clothing. & + \\
        \hline
        \IF{} & an unpretentious sexy atmosphere lends itself to the above average wine-list and a \textcolor{blue}{menu} that can stand-up to any other restaurant ... & + \\
        \hline
    \end{tabular}
    \caption{Showcasing Top-1 Explanations. Aspect terms are in blue, and the spans are in bold font. \tracinf{} do not highlight either training or testing span; \tracinp{} highlights training span; \tracinpp{} highlights both training and test spans. \tracinpp{} and \IFpp{} can help users understand which span of $z$ influenced which span of $z'$, which \tracinf{} and \IF{} do not provide.}
    \label{tab:cases}
\end{table*}

%\section{Discussions}
%\label{sec:disc}

%\paragraph{Application of \tracinpp{}} \zhu{this part does not seem to provide a clear and complete message. or maybe we move this to future work. }The \tracinpp{} is generally applicable to NLP problems. Although we pre-annotate spans for data, it may not be necessary for cases such as long documents where a single sentence can conveniently be an explanation unit. The step can be 1) using \texttt{imp}() function to identify the important sentence(s), and 2) apply \tracinpp{} on each sentence individually, and then combine the lists. Our method can also be applied to text generation models since the importance of a span can be conveniently determined by input masking.
%\paragraph{Method Connections}
%\zhu{I think it's better to discuss the connections right after you introduce TracInF.}\IF{} and \tracin{} has very similar formation. The difference is on how the variance of gradient estimates are reduced: The inverse-Hessian of \IFp{} can be seen as regularizing the gradient of $z$ and $z'$ by the co-variance of $\hw$ (Eq. \ref{eq:IF}); whereas \tracin{}'s checkpoint averaging can be seen as another way to reduce this variance (Eq. \ref{eq:tracin}). The proposed \tracinp{} and \tracinpp{} methods are essentially breaking the temporal dependencies among the checkpoints in vanilla \tracin{}, and it may lead to more robust gradient estimates, which effect has been shown empirically in \textcolor{red}{model ensembles \cite{} adversarial training \cite{}, reinforcement learning \cite{} and stochastic optimizations \cite{}.}

\section{Related Work}
Popular explanation methods include gradient-based \cite{axiomatic_attribution_2017_Ankur}, attention-based \cite{Clark-Manning-2019,attention_not_exp_Sarthak2019,attention_not_not_exp_Sarah}, as well as sample-based \cite{koh2017understanding,yeh2018nips,pruthi2020estimating} methods.
\paragraph{Major Progress on Sample-based Explanation Methods}
There have been a series of recent efforts to explain black-box deep neural nets (DNN), such as LIME \cite{lime-RibeiroSG16} that approximates the behavior of DNN with an interpretable model learned from local samples around prediction, Influence Functions \cite{koh2017understanding, koh2019} that picks training samples as explanation via its impact on the overall loss, and Exemplar Points \cite{yeh2018nips} that can assign weights to training samples. \tracin{} \cite{pruthi2020estimating} is the latest breakthrough that overcomes the computational bottleneck of Influence Functions with the cost of faithfulness.

\paragraph{The Discussion of Explanation Faithfulness in NLP} The issue of Faithfulness of Explanations was primarily discussed under the explanation generation context \cite{camburu2018nips} where there is no guarantee that a generated explanation would be faithful to a model's inner-workings \cite{jacovi-goldberg-2020-towards}. In this work, we discuss faithfulness in the sample-based explanations framework. The faithfulness to model either can be guaranteed only in theory but not in practice \cite{koh2017understanding} or can not be guaranteed at all \cite{pruthi2020estimating}.

\paragraph{Sample-based explanation methods for NLP}
Han et al. \shortcite{han2020explaining} applied \IF{} for sentiment analysis and natural language inference and also studied its utility on detecting data artefacts \cite{gururangan-etal-2019-variational}. Yang et al. \shortcite{yang-etal-2020-generative} used Influence Functions to filter the generated texts. The one closest to our work is \cite{meng-etal-2020-structure} where a single word is used as the explanation unit. Their formation uses gradient-based methods for single words, while ours can be applied to any text unit granularity using text masking.

\paragraph{Explanation of NLP Models by Input Erasure}
Input erasure has been a popular trick for measuring input impact for NLP models by replacing input by zero vector \cite{li2016understanding} or by marginalization of all possible candidate tokens \cite{kim-etal-2020-interpretation} that arguably dealt with the out of distribution issue introduced by using zero as input mask. Similar to \cite{kim-etal-2020-interpretation, li2020bert, jacovi-goldberg-2020-aligning} we also use ``[MASK]'' token, with the difference that we allow masking of arbitrary length of an input sequence. 

\paragraph{Evaluations of Sample-based Methods}
A benchmark of evaluating sample-based explanation methods has not been agreed upon. For diagnostic purposes, Koh et al. \shortcite{koh2017understanding}  proposed a self-explanation method that uses the training instances to explain themselves; Hanawa et al. \shortcite{2020Evaluation} proposed the label and instance consistency as a way of model sanity check. On the non-diagnostic setting, sample removal and re-training \cite{han2020explaining,hooker2018benchmark} assumes that removing useful training instances can cause significant accuracy loss; input enhancement method assumes useful explanations can also improve model's decision making at model input side \cite{hao-2020-evaluating}, and manual inspections \cite{han2020explaining, meng-etal-2020-structure} were also used to examine if the meanings of explanations align with that of the test instance. In this paper, we automate this semantic examination using the embedding similarities.

\section{Future Work}
\tracinpp{} opens some new questions: 1) how can we generalize \tracinpp{} to cases where test spans are unknown? 2) Can we understand the connection between \IF{} and \tracin{} which may spark discoveries on sample-based explanation methods? 3) How can we apply \tracinpp{} to understand sequence generation models? 

\section*{Acknowledgement}
This work is supported by the MIT-IBM Watson AI Lab. The views and conclusions are those of the authors and should not be interpreted as representing the official policies of the funding agencies. We thank anonymous reviewers for their valuable feedback. We also thank your family for the support during this special time.

\bibliographystyle{acl_natbib}
\bibliography{anthology,acl2021}

%\appendix

\onecolumn

\appendix

\section{Span extraction details}
The model we apply the huggingface \cite{Wolf2019HuggingFacesTS} pre-trained RC model ``phiyodr/roberta-large-finetuned-squad2'' \cite{rc-squad2.0} which is chosen based on our comparison to a set of similar models on SQuAD 2.0 dataset. We use the SQuAD 2.0-trained model instead of 1.0 because the data is more challenging since it involves multiple passages, and the model has to compare valid and invalid passages for answer span extraction, a case similar to the dataset we use.
Templates we used are:
\begin{table}[htb]
    \centering
    \begin{tabular}{|l|}
    \hline
         How is the X? \\
         How was the X? \\
         How are the X? \\
         How were the X? \\
         How do you rate the X? \\
         How would you rate the X? \\
         How do you think of the X? \\
         What do you think about the X?\\
         What do you say about the X? \\
         What happened to the X? \\
         What did the X do? \\
         \hline
    \end{tabular}
    \caption{Templates for RC model}
    \label{tab:my_label}
\end{table}
The heuristics when the RC model fails: 1) We consider RC model fails when no span is extracted, or the entire text is returned as an answer. 2) We identify the location of the term in the text and expand the scope from the location both on the left and on the right, and when sentence boundary is found, we stop and return the span as the span for the term. Note that we do find cases where the words around a term do not necessarily talk about the term. However, we found such a case to be extremely rare.

\section{Derivation of \IFpp{}}
\begin{align*}
    \mathcal{I}_{\text{pert,loss}} & (X_{ij},z_{-kl}; \hat{\theta}) \\ 
    &:=\nabla_{\epsilon} \text{imp}(X_{ij}| X ; \hat{\theta}_{\epsilon, z_{-ij}, z} ) \Bigg |_{\epsilon=0}\\
    &= \frac{d \text{imp}(X_{ij}| X ; \hat{\theta})}{d \hat{\theta}}  (\frac{d \hat{\theta}_{\epsilon, z_{-kl}, z}}{d\epsilon }\Bigg|_{\epsilon=0})\\
    &= ( \nabla_{{\theta}} O_y(X, \hat{\theta}) - \nabla_{{\theta}} O_y(X_{-ij}, \hat{\theta})) (\frac{d \hat{\theta}_{\epsilon, z_{-kl}, z}}{d\epsilon } \Bigg |_{\epsilon=0}) \\
    &= - ( \nabla_{{\theta}} O_y(X, \hat{\theta}) - \nabla_{{\theta}} O_y(X_{-ij}, \hat{\theta})) H_{\hat{\theta}}^{-1}(\nabla_{\theta} L(z_{-kl}, \hat{\theta}) -\nabla_{\theta} L(z, \hat{\theta}) ) 
\end{align*}

\section{Derivation of \tracinp{} and \tracinpp{}}

Similar to \IF \cite{koh2017understanding} and \tracin \cite{pruthi2020estimating}, we start from the Taylor expansion on point $\hw_t$ around $z'$ and $z'_{-ij}$ as
\begin{align*}
\mathcal{L}(\hw_{t+1}, z') \sim \mathcal{L}(\hw_{t}, z') + \nabla \mathcal{L}(\hw_t, z') (\hw_{t+1} -  \hw_t) \\
\mathcal{L}(\hw_{t+1}, z'_{-ij}) \sim \mathcal{L}(\hw_{t}, z'_{-ij}) + \nabla \mathcal{L} (\hw_t, z'_{-ij}) (\hw_{t+1} - \hw_t) 
\end{align*}
If SGD is assumed for optimization for simplicity, $(\hw_{t+1} - \hw_t)=\lambda \nabla \mathcal{L}(\hw_t, z)$. Thus,  putting it in above equations and perform subtraction, we obtain 
%If we define $\Delta\hw = (\hw_{t+1} -  \hw_t)$
\begin{align*}
\mathcal{L}(\hw_{t+1}, z') - \mathcal{L}(\hw_{t+1}, z'_{-ij}) \sim \mathcal{L}(\hw_{t}, z'_{-ij}) - \mathcal{L}(\hw_{t}, z') + [ \nabla \mathcal{L}(\hw_t, z') - \nabla \mathcal{L}(\hw_t, z'_{-ij})] \lambda \nabla \mathcal{L}(\hw_t, z)
\end{align*}

And, 

\begin{align*}
\text{imp}(x'_{ij} | z'; \hw_{t+1}) - \text{imp}(x'_{ij} | z'; \hw_{t})  \sim [ \nabla \mathcal{L}(\hw_t, z'_{-ij}) -\nabla \mathcal{L}(\hw_t, z') ]  \lambda \nabla \mathcal{L}(\hw_t, z)
\end{align*}

So, the left term is the change of importance by parameter change; we can interpret it as the change of importance score of span $x_{ij}$ w.r.t the parameter of networks. Then, we integrate over all the contributions from different points in the training process and obtain

\begin{align*}
\tracinp{}(x'_{ij} | z', z)  = \sum_t [ \nabla \mathcal{L}(\hw_t, z'_{-ij}) -\nabla \mathcal{L}(\hw_t, z') ]  \lambda \nabla \mathcal{L}(\hw_t, z)
\end{align*}
The above formation is very similar to \tracin where a single training instance $z$ is evaluated as a whole. But we are interested in the case where an meaning unit $x_{kl}$ in $z$ can be evaluated for influence. Thus, we apply the same logic of the above equation to $z_{-kl}$, the perturbed training instance where token $k$ to $l$ is masked, as
\begin{align*}
\tracinp{}(x'_{ij} | z', z_{-kl})  = \sum_t [ \nabla \mathcal{L}(\hw_t, z'_{-ij}) -\nabla \mathcal{L}(\hw_t, z') ]  \lambda \nabla \mathcal{L}(\hw_t, z_{-kl})
\end{align*}
Then, the difference  $\tracinp{}(x'_{ij} | z', z) - \tracinp{}(x'_{ij} | z', z_{-kl})$ can indicate how much impact a training span $x_{kl}$ on test span $x'_{ij}$.  Formally, the influence of $x_{kl}$ on $x'_{ij}$ is 
\begin{align*}
\tracinpp{}(x'_{ij} , x_{-kl}| z', z)  = \lambda \sum_t [ \nabla \mathcal{L}(\hw_t, z'_{-ij}) -\nabla \mathcal{L}(\hw_t, z') ]   [\nabla \mathcal{L}(\hw_t, z) - \nabla \mathcal{L}(\hw_t, z_{-kl})]
\end{align*}
We denote that such a form is very easy to implement, since each item in summation requires only four (4) gradient estimates.

\end{document}